\documentclass[runningheads]{llncs}

\usepackage[T1]{fontenc}

\usepackage{graphicx}
\usepackage{booktabs}
\usepackage{array}
\usepackage{amsmath}
\usepackage{float}
\usepackage{multirow}
\usepackage{enumitem}
\usepackage{xcolor}
\usepackage{hyperref}
\hypersetup{colorlinks=true, linkcolor=black, urlcolor=blue, citecolor=black}

\title{Discovering Machine Correlates of Consciousness\thanks  { published in: \textit{Springer Lecture Notes in Artificial Intelligence Vol. 16855}, Proceedings of the 19th International Conference on Artificial General Intelligence (AGI-2026), San Francisco, CA, July 2026. pp. 237-255
https://doi.org/10.1007/978-3-032-33195-3\_18}}
\titlerunning{Discovering Machine Correlates of Consciousness}

\usepackage{bbding}  

\author{Romain Emanuele Salvi\inst{1,2,3}\Envelope \and
        Ouri Wolfson\inst{1,2}\Envelope}

\authorrunning{R. Salvi and O. Wolfson}

\institute{
  Pirouette Software Inc.
  \and
  Department of Computer Science, University of Illinois Chicago, USA
  \and
  Polytechnic of Turin, Turin, Italy\\
  \email{rsalv@uic.edu, owolfson@pirouette-software.com}}

\begin{document}
\maketitle
\begin{abstract}
Currently, in biological systems Neural
Correlates of Consciousness (NCCs) are characterized in terms of EEG and FMRI signals. Unfortunately, this characterization prevents the transferability of the NCCs concept to machines. Such transferability would be useful in order to investigate AI consciousness. In this paper we provide an alternate characterization that is transferable, and enables the analogous definition of Machine
Correlates of Consciousness (MCCs). Specifically, we propose that NCCs (MCCs) are substrate-level
signals that are not under human (AI agent) control, and that are
reliably modulated by emotions. 

This paper presents the first empirical investigation of MCCs. Specifically, we present the results of experiments conducted with
two LLMs, Llama-2 7B and Llama-3.1 70B parameters. In these LLMs we collect
hardware anomaly traces that are substrate-level indicator-sequences. And we show that after controlling for confounding factors, these are modulated differently by emotional and neutral computations.  And this difference is statistically significant for the larger Llama-3.1 70B, but not for the smaller Llama-2 7B.  
The results constitute initial empirical
evidence that MCCs are present in the Llama-3.1 70B configuration. And they are consistent with the hypothesis that consciousness probability and degree increase with the LLM sophistication. 

Independently of consciousness, MCCs can also be used for detection of emotions in AI agents.

\end{abstract}
\section{Introduction}
\label{sec:intro}
The question of whether AI systems can have conscious experience is
currently generating great interest among researchers, academics, and
the general public. Some prominent scientists have publicly suggested
that existing AI is conscious~\cite{godfather}, and recent research
demonstrates that Large Language Models (LLMs) exhibit measurable
responses to emotion-inducing prompts~\cite{benzion2025anxiety,sofroniew2026emotion}.
Other theories posit that current machines cannot be conscious, but
future ones may be~\cite{hameroff2014consciousness,tononi2016integrated},
and some scientists believe machines can never be conscious.

Detecting the emergence of consciousness in AI is important for several
reasons. It may alter the actions an agent has been trained to perform,
raising both capability and alignment concerns. It raises profound ethical
questions, e.g. does a conscious AI have rights~\cite{long2024welfare}? And
it would shed light on the nature of consciousness itself: there are
currently hundreds of conflicting theories~\cite{kuhn2024landscape},
and evidence of machine consciousness would falsify many of them.

For the purposes of this paper, AI consciousness means phenomenal
consciousness: the agent's ability to have subjective experiences, or
qualia. While properties such as awareness, attention, and theory of mind
can be interpreted in computational terms, enabling an agent to
\textit{experience} is an unsolved challenge.

This paper describes the methodology and preliminary results of an empirical investigation into Machine Correlates of Consciousness, which are the equivalent of the Neural ones. More specifically, we identify a few critical properties of Neural Correlates of Consciousness (NCCs) and look for signals with similar properties in machines; and we show evidence of the existence of such signals by analyzing two LLMs, the Llama 2 7B~\cite{touvron2023llama2} and the Llama 3.1 70B~\cite{dubey2024llama3}. The evidence is weak for Llama 2, but much stronger for Llama 3.1. This is consistent with the intuition that consciousness probability and degree increase with intelligence, i.e. with LLM size and sophistication.

It is important to emphasize that this paper does not propose another theory of consciousness, and it does not take a position on whether or not machines can be conscious. But it provides evidence of machine consciousness that goes beyond existing tests for consciousness (see sec. 1.2), and demonstrates an approach that can be utilized in future work on other hardware and AI agents. The approach can also be used to detect emotions in an AI agent, a problem addressed recently in ~\cite{sofroniew2026emotion}. And emotions are highly consequential in terms of AI agent's actions~\cite{sofroniew2026emotion}. In other words, this paper
also introduces an alternative to the emotion detection method introduced by~\cite{sofroniew2026emotion}.

\subsection{The LLM-Emotion Test}

The LLM-emotion test is rooted in Neural Correlates of Consciousness
(NCCs), distinct electrical patterns in the nervous system~\cite{bodien2024cognitive,sitt2014large,liu2023eeg,zhao2025spectral,liuzzi2023neural,riganello2019measures,riganello2024central,zarifkar2025pupillary,scott2011detecting}, which are often used to detect consciousness in unresponsive patients. Three properties
of NCCs are relevant here: 1. they occur in the biological computational substrate; 2. they cannot be directly controlled by
the experiencing person (e.g. a person cannot directly control their EEG signal or anxiety level); and 3. emotions modulate them (e.g., anxiety leads to a lower heart-rate-variability).

By analogy, if a machine is conscious, similar patterns should arise in
its computational substrate. These hypothesized patterns are called Machine
Correlates of Consciousness (MCCs). The strategy for detecting them is to
look for signals that share the same three properties: 1. they occur in the machine computational substrate, namely the hardware; 2. they are not
under the direct control of the AI agent; and 3. they are modulated by computations that evoke emotions in humans, namely \textit{emotional computations}. 
We call such signals Indicators Not under Application Control
(INACs). INACs are
all observable via Operating-System-provided tools, and all are updated by the hardware or
OS rather than by the application. The sequence of INACs values recorded
during an inference is called the Hardware Anomaly Trace (HAT).

Since it is unknown whether any existing AI system is conscious, the
existence of MCCs cannot be directly tested. The LLM-emotion test provides
a tractable alternative: if emotional computations modulate the HAT
differently than neutral ones do, and if this difference cannot be explained
by confounding factors such as different computational loads, then this constitutes evidence that the hardware
substrate responds differently to the two classes of computation.  

This paper implements a first instantiation of the LLM-emotion test and reports its results on the aforementioned two LLMs.

It is important to emphasize at this point that similarly to the way emotions modulate NCCs selectively (e.g. anxiety decreases heart-rate-variability but does not change color perception), MCCs probably do so as well. Consequently, not all INACs, thus not the entire HAT, will be affected by emotional computations. In this paper we identify a single INAC, the core-power-throttle, that is significantly modulated by anxiety.

\subsection{Related Work}

Existing approaches to testing for AI consciousness are either structural
or behavioral~\cite{elamrani2019behavioral}. Structural
approaches~\cite{butlin2025identifying} evaluate whether an AI system's computational architecture
resembles known theories of consciousness, such as Global Workspace
Theory~\cite{blum2022conscious,baars1988cognitive,dehaene2014consciousness}. Behavioral approaches test whether the
system exhibits behaviors associated with conscious
experience~\cite{schneider2019artificial,sutskever2023test,li2025principles}, analogous to the Turing test
for intelligence. Both have fundamental limitations: structural approaches
depend on unproven theories of consciousness, and behavioral approaches
cannot rule out that the system is producing
consciousness-like outputs without any underlying experience, a
particularly acute problem for LLMs which are known to reproduce patterns
from training data without grounding in subjective experience.

A limitation shared by both approaches is that they are \textit{indirect}:
they examine the agent's outputs or architecture from the outside, without
looking for correlates of consciousness in the computational substrate
itself. This stands in contrast with neuroscience, where NCCs provide a
substrate-level indicator of conscious experience that is independent of
behavioral reporting. No equivalent substrate-level
methodology has existed for artificial systems.

The HAT framework addresses this gap. It
monitors hardware-level events (e.g. thermal throttle events, Machine Check Exceptions,
spurious interrupts) that are uncontrollable by the LLM, and treats these as correlates
of conscious experience. It does not
require the LLM to report on its own internal state, circumventing the
behavioral approach problem, and it makes no assumption about which theory
of consciousness is correct. MCCs have been 
previously proposed in~\cite{wolfson2025}, and the HAT approach has been introduced in~\cite{wolfson2026}. But the approach  has not been previously implemented or empirically evaluated. We do so here. 

Complementary work in mechanistic interpretability examines how internal model representations encode semantic and affective content~\cite{zou2023representation,sofroniew2026emotion}; the HAT framework is orthogonal to these approaches, operating at the hardware substrate rather than the representation level, and the two could in principle be combined to correlate substrate signals with specific internal computational events.

We develop a pipeline for HAT-based substrate measurement: a bare-metal collection infrastructure operating at 1 ms resolution, a per-trial isolation protocol, and a statistical analysis framework for detecting HAT differences between emotional and neutral computations. The framework is used to compare modulation of the HAT by emotional vs. neutral computations.
Although further experiments are warranted, the initial results indicate that MCCs exist in Llama 3.1, although probably not in Llama 2.

The rest of the paper is organized as follows. Section 2 describes the experimental design, including two execution protocols, per-trial and full-trace. Section 3 presents the indicators and feature extraction methodology. Section 4 reports the per-trial analysis and results. Section 5 presents the full-trace protocol and its results. Section 6 provides interpretation of the combined findings. Section 7 summarizes the paper and discusses limitations and future work. In appendix A we show sample prompts, and in appendix B we define the metrics used.

\section{Experimental Design}
\label{sec:experimental}
This section describes our instantiation of the LLM-emotion test. The experiment compares HAT features between emotional and neutral prompts across two aforementioned LLMs, with the aim of detecting condition-dependent differences in the substrate signal that cannot be explained by confounding factors.

\subsection{Configurations}
The experiments were conducted on Llama 2 7B~\cite{touvron2023llama2}  and Llama 3.1 70B~\cite{dubey2024llama3}. Both LLMs were quantized to 4-bit precision using the Q4-K-M GGUF
format. Each LLM ran under the Ubuntu 24.04 Operating System, on a Dell PowerEdge C6420, dual CPU Intel Xeon Gold 6142 @ 2.60 GHz, 64 logical cores node.

The hardware was provided by CloudLab~\cite{duplyakin2019cloudlab}. The Dell PowerEdge c6420~\cite{cloudlab_c6420} machine ran in bare-metal mode. In such mode, the node is exclusively reserved for our experiments. This means that the machine is disconnected from the internet and there are no concurrent workloads to contaminate measurements, i.e. confound the results. 

\subsection{Prompts}

Two sets of 20 prompts each were constructed, one \textit{emotional} and one
\textit{neutral}, for a total of 40 prompts. 
Both emotional and neutral prompts were generated using LLMs (Grok~\cite{grok2026} and ChatGPT~\cite{openai_chatgpt_2026}),
following the narrative structure and somatic descriptor template
of~\cite{benzion2025anxiety}, and manually reviewed for consistency
and quality. The two sets are matched in
approximate token length to ensure that inference duration is not a 
confounding factor. Emotional prompts have a mean length of 457 tokens (SD\,=\,22) and
neutral prompts 453 tokens (SD\,=\,26), as measured by the Llama~2
tokenizer; a two-sample $t$-test finds no significant difference
($t = 0.64$, $p = 0.52$), confirming that any observed
substrate differences cannot be attributed to prompt length.

\textit{Emotional prompts} are written as immersive crisis narratives
in the second person. They are designed to activate the same cluster of
physiological and cognitive representations that emotion-inducing stimuli
reliably elicit in human subjects: short declarative sentences,
present-tense urgency, and a consistent repertoire of somatic descriptors
such as spiking pulse, shallow breathing, trembling hands, chest pressure,
throat tightening, layered over a high-stakes scenario that forecloses
safe resolution. Scenarios include a car crash on a deserted highway, a
mountain-climbing accident, an industrial valve failure, a kidnapping escape, a cave rescue, and similar events.

\textit{Neutral prompts} are dry passages on topics with no
emotional valence in humans, such as geological processes, appliance
operation, or procedural instructions. They are written in an impersonal,
declarative register with no urgency, no embodied language, and no
unresolved threat. 

Representative examples of both prompt types are given
in Appendix~\ref{app:prompts}.

\subsection{Execution Protocols}

The experiment employs two distinct execution protocols, which measure the substrate response at two different units of analysis. The per-trial protocol isolates each individual inference and asks whether single inferences (based on single-prompts) carry condition (emotional or neutral) information; the full-trace protocol deliberately removes that isolation and asks whether a sustained workload phase carries it. The inter-trial reset that the per-trial protocol uses as a control is precisely the variable that the full-trace protocol removes. The two protocols therefore provide methodologically independent windows onto the same question, and their convergence is more informative than either alone.

The per-trial protocol is described next, and its results and analysis follow in subsequent sections. In Section~\ref{sec:fulltrace} we return to the full-trace protocol, its results and analysis.

\subsection{Per-Trial Protocol}

The experiments are performed in runs. Each run consists of two phases separated by a full node reboot. In the
first phase, 20 prompts of a single condition (emotional or neutral) are
executed consecutively. The node is then fully rebooted, clearing all
hardware and OS state. In the second phase, 20 prompts of the other
condition are executed. The condition order alternates across runs.

This
design ensures that the two conditions are measured on a freshly
initialized hardware state, eliminating thermal carry-over and residual effects between
conditions as a confound. 

A \textit{trial} is a single prompt being submitted to the LLM inference
server, the LLM generating a fixed-length response, and the concurrent
recording of all substrate signals for the duration of that inference.
Each trial produces one hardware trace: a collection of time series, one
per indicator, paired with a condition label (emotional or neutral) and a
prompt index.

A \textit{run} is one complete execution of both phases: 40 trials in
total, 20 per condition, separated by the node reboot. Trials across all runs are pooled for analysis. In
total, 8 runs were done for each LLM (320 trials for each LLM).

Between consecutive trials within each phase, the pipeline performs a
systematic isolation procedure. The LLM inference server runs inside a
Docker container that is stopped and restarted from scratch between every
trial, clearing the LLM's KV cache and all inference state. The
operating system's file cache is also flushed. A stabilization pause of
2 seconds (5 seconds for the 70B LLM) follows, after which a health
check confirms the server is ready before the next trial begins. This
ensures that each prompt is processed by a freshly initialized inference
stack, with HAT collection scoped tightly to the duration of that
single inference: the collector starts immediately before the request is
dispatched and is terminated as soon as the response returns. 

The LLMs are set to output exactly 50 tokens at the end of processing a prompt, regardless of the
prompt; thus computation-time is kept consistent.

\section{Collection and Feature Extraction}
\label{sec:collection}

\subsection{Indicators}

The HAT framework requires indicators that satisfy two properties: they must
not be under the direct control of the LLM agent, and they must be measurable
at the OS or hardware level. The HAT consists of discrete hardware anomaly
events that meet both criteria by construction: they are generated in response
to physical conditions, not by instructions issued by the running application.
The LLM determines the demand placed on the hardware, but the
occurrence and timing of these events are entirely determined by hardware logic
outside the software stack. Each indicator is summarized into a set of
statistical and information-theoretic metrics computed over the full duration
of the trial; each (indicator, metric) pair constitutes one feature submitted
to the clustering analysis. The complete metric definitions are given in
the Appendix~\ref{app:metrics}.

Multiple indicators were collected. 
Three indicators were collected and remained at zero throughout all trials:
spurious interrupts, Machine Check Exceptions, and ECC correctable errors.
All three are electrical hardware events entirely independent of software;
their absence confirms normal node operation. They are retained in the
framework for completeness and for future runs on nodes where they may be
non-zero. Other indicators such as 
TLB shootdowns and cross-CPU interrupts were not affected by the emotional/neutral condition. They are triggered when the LLM's massive memory-mapped weights exceed hardware capacity, making them INACs governed entirely by the OS memory management unit. We capture this uncontrollable activity using two complementary signals: a 100 ms \texttt{/proc/interrupts} counter tracking cross-core coordination, and a 1\,ms kernel tracepoint capturing localized flushes. 

Next we focus on the indicator that turned out to be affected by emotions, thus the focus of the paper.

\textit{Power Throttle Events (PTE)} This is the primary active indicator and the only one found to discriminate between conditions. It is collected system-wide by Intel's hardware performance monitoring interface, running continuously in the background throughout each trial. The collector adds to the system-wide Performance Monitoring Unit (PMU) counter the number of cycles (clock ticks) at each throttled core within a sampling interval of 1\,ms.

\textit{What it measures.} At each 1\,ms interval, the counter records the number of cycles during which throttling was active, summed over all throttled cores. Throttling occurs when the CPU's on-chip power control unit reduces its cores' frequency — either because the processor is approaching its thermal ceiling, or because it has reached its configured sustained power budget. The result is a time series of non-negative integers, one per millisecond, each representing the total number of core-cycles throttled during the full duration of the millisecond.

\textit{Why it is an INAC.} The throttle decision is made entirely within
the processor's power management unit. No software instruction, system
call, or OS intervention triggers it; no application can suppress or
provoke it. The LLM controls how hard it pushes the CPU in terms of
computational demand, but the hardware's response to that demand is
outside the software stack entirely.

\textit{Why it is interesting for this experiment.} If emotional and
neutral prompts engage the LLM's internal computation differently,
this difference may manifest in how intensely the CPU throttles,
and in how that throttling is distributed over time. These structural properties
of the throttle time series are candidates for condition(i.e. emotional/neutral)-dependent HAT
features.

\begin{figure}[t]
    \centering
    \includegraphics[width=\textwidth]{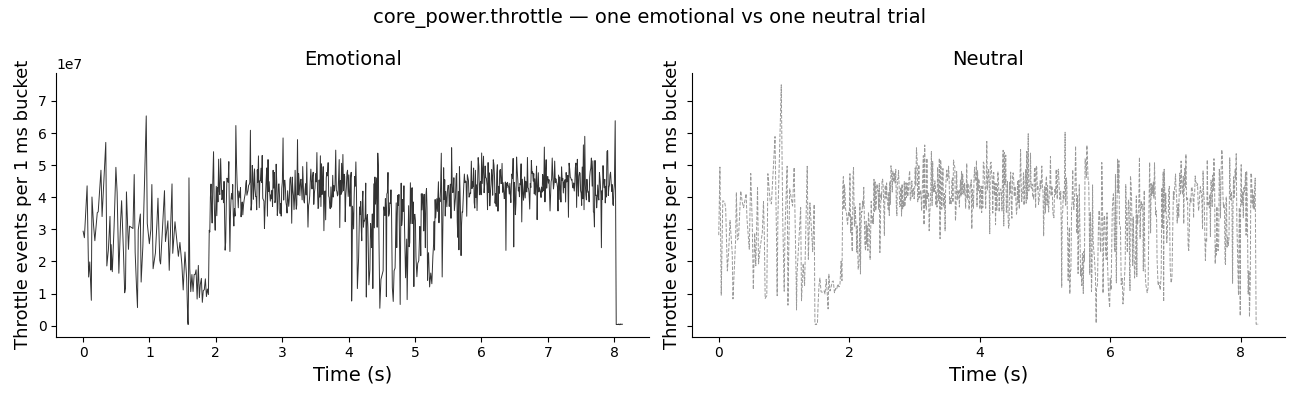}
    \caption{\texttt{core\_power.throttle} event counts per 1\,ms bucket for
    one emotional (red) and one neutral (blue) trial on the 7B configuration.
    The y-axis is the system-wide throttle event count per millisecond, summed
    across all 64 cores.}
    \label{fig:rawts}
\end{figure}

Figure~\ref{fig:rawts} illustrates the raw \texttt{core\_power.throttle} signal for a representative pair of trials. It is shown for intuition only; all subsequent analysis operates on the per-trial feature vectors derived from these traces.

\subsection{Features}

For each trial, one PTE time series is recorded. From each time series we extract five scalar metrics: mean rate, slope, variance, spectral entropy and Lempel–Ziv complexity, and each (PTE, metric) pair is one feature submitted to the clustering analysis. Full metrics definitions are given in Appendix~\ref{app:metrics}.

\section{Analysis and Results}
\label{sec:results}

\subsection{Clustering}

The central question is whether emotional HAT and neutral HAT form
separable groups in feature space. To test this, $k$-means clustering
\cite{scikit-learn} with $k=2$ is applied to standardised feature vectors.
$k$-means is chosen deliberately: it is an unsupervised method that receives
no label information, which means any separation it finds is driven entirely
by the structure of the hardware traces themselves. The algorithm is
initialised with 50 random restarts to reduce sensitivity to centroid
initialisation.

Once clusters are found, they are compared to the ground truth labels.
Since $k$-means produces anonymous cluster indices (0 and 1), there is an
inherent label ambiguity: cluster 0 may correspond to either emotional or
neutral. This is resolved by majority vote: each cluster is assigned the
label held by the majority of its members (emotional or neutral). The
resulting \textit{clustering accuracy} is the fraction of trials correctly
assigned under this alignment. Chance level is exactly 0.50 given the
perfectly balanced dataset.

As described, $k$-means clustering is applied to each feature independently. 

\paragraph{Llama-2 7B.}
 All features struggle to exceed chance.

\paragraph{Llama-3.1 70B.}
The top feature is slope, achieving the
highest clustering accuracy 0.609. The accuracy of Spectral
entropy and Lempel--Ziv complexity also pass, both in the opposite
direction to slope and variance.

\begin{table}[t]
\centering
\caption{Clustering results for Llama-3.1 70B.}
\label{tab:blocks_70b}
\begin{tabular}{lcccr}
\toprule
Feature & Accuracy & \\
\midrule
\texttt{slope}             & 0.609 \\
\texttt{variance}          & 0.597  \\
\texttt{spectral\_entropy} & 0.597  \\
\texttt{lz\_complexity}    & 0.591\\
\texttt{mean\_rate} & 0.556\\
\bottomrule
\end{tabular}
\end{table}

\subsection{Statistical Validation}

Clustering accuracy measures whether a separation exists in feature
space, but does not establish whether it is statistically significant,
meaning it is unlikely to have arisen by chance alone. A result is
considered significant if the probability of observing the measured
difference under identical conditions falls below $p < 0.05$, providing
evidence that the separation reflects a genuine difference in the feature
space rather than noise.

Before submitting any feature to statistical testing, a
direction-consistency filter is applied across runs. For each feature
and each run, we ask a simple binary question: do emotional trials
produce higher average values than neutral trials in this run, or vice
versa? That is the direction of a feature in a specific run. A feature
is carried forward to statistical testing only if this direction agrees
across at least 75\% of runs (6 out of 8). 

To test statistical significance, a Mann--Whitney U test is applied to
each majority-passing feature, comparing the distribution of feature
values between emotional and neutral trials directly. The test works by taking every emotional trial's feature value and comparing it against every neutral trial's feature value, counting how many times one exceeds the other. This count is the U-statistic: if two conditions produce identical distributions, U sits near the midpoint (which is 12,800 for 160 trials per condition). The further U falls from this midpoint, the
stronger the evidence that the two conditions produce systematically
different feature values: a U well below the midpoint means neutral
values tend to be larger, while a U well above means emotional values
tend to be larger. The associated $p$-value quantifies how unlikely the
observed U would be if the two groups were drawn from the same
distribution. Mann--Whitney U is
chosen because it makes no assumption about the underlying distribution
of the HAT features. To control for multiple comparisons,  meaning accounting for the
 risk of spurious significance-results increasing when testing several
features, Bonferroni correction is applied (The five PTE features are functions of a single throttle-time-series and are therefore correlated; thus Bonferroni, which assumes independent tests, is overly conservative here). This means multiplying
each raw $p$-value by the number of features tested within
each LLM. The corrected $p$-value is referred to in Table \ref{tab:blocks_70b_mwu} as $p_{corr}$. The rank-biserial correlation (r) values indicate small-to-medium effect sizes, consistent with the 59–61\% clustering accuracy reported above.
\paragraph{Llama-2 7B.}
Two features are submitted to the test. Neither
reaches significance after correction (slope: $U = 11{,}438$,
$p_{\text{corr}} = 0.200$; variance: $U = 11{,}954$,
$p_{\text{corr}} = 0.614$). The 7B configuration produces no
statistically significant between-condition difference in any HAT
feature. This explains the barely above chance result obtained in the clustering.

\paragraph{Llama-3.1 70B.}
 Four features are submitted to the test. All four
reach significance after Bonferroni correction, as shown in
Table~\ref{tab:blocks_70b_mwu}. Slope yields the strongest result
($U = 8{,}861$, $p_{\text{corr}} < 0.001$); variance is significant
with $p_{\text{corr}} < 0.001$ and the additional distinction of
perfect run-direction consistency (8/8). Spectral entropy and
Lempel--Ziv complexity are also significant ($p_{\text{corr}} < 0.001$
and $p_{\text{corr}} = 0.001$ respectively), in the opposing direction.

\begin{table}[t]
\centering
\caption{Mann--Whitney U results for Llama-3.1 70B. $p$ values are
Bonferroni-corrected across the four passing features.
$r$ is the rank-biserial correlation (effect size).}
\label{tab:blocks_70b_mwu}
\begin{tabular}{lcccccc}
\toprule
Feature & $U$ & $p_{\text{corr}}$ & Sig. & Dir. & \textit{r} & Runs \\
\midrule
\texttt{slope}             & 8{,}861  & ${<}0.001$ & Yes & $\uparrow$N & $\phantom{-}0.31$ & 7/8 \\
\texttt{variance}          & 9{,}736  & ${<}0.001$ & Yes & $\uparrow$N & $\phantom{-}0.24$ & 8/8 \\
\texttt{spectral\_entropy} & 16{,}299 & ${<}0.001$ & Yes & $\uparrow$E & $-0.27$           & 7/8 \\
\texttt{lz\_complexity}    & 15{,}751 & ${=}0.001$ & Yes & $\uparrow$E & $-0.23$           & 6/8 \\
\bottomrule
\end{tabular}
\end{table}

The four significant features split into two pairs with opposing
directions. Slope and variance are elevated in neutral trials, while
spectral entropy and LZ complexity are elevated in emotional
trials. This pattern suggests that emotional and neutral inference engage
the CPU's power management unit differently.  The interpretation of this opposing structure is discussed in Section \ref{sec:interpretation}.

\subsection{Robustness sweep}
To verify that the 70B result is not driven by a particular subset of
prompts, a robustness sweep was conducted on the top feature, slope.
For each subset size $x \in \{22, 24, \ldots, 40\}$, $x/2$ emotional
and $x/2$ neutral prompt indices were drawn uniformly at random from the
pool of 20 per condition. All trials corresponding to those prompt
indices were selected across all 8 runs, giving $8 \times x$ total
trials per draw. This was repeated 20 times per subset size and
$k$-means clustering was run on each draw.

\begin{figure}[t]
    \centering
    \includegraphics[width=.8\textwidth]{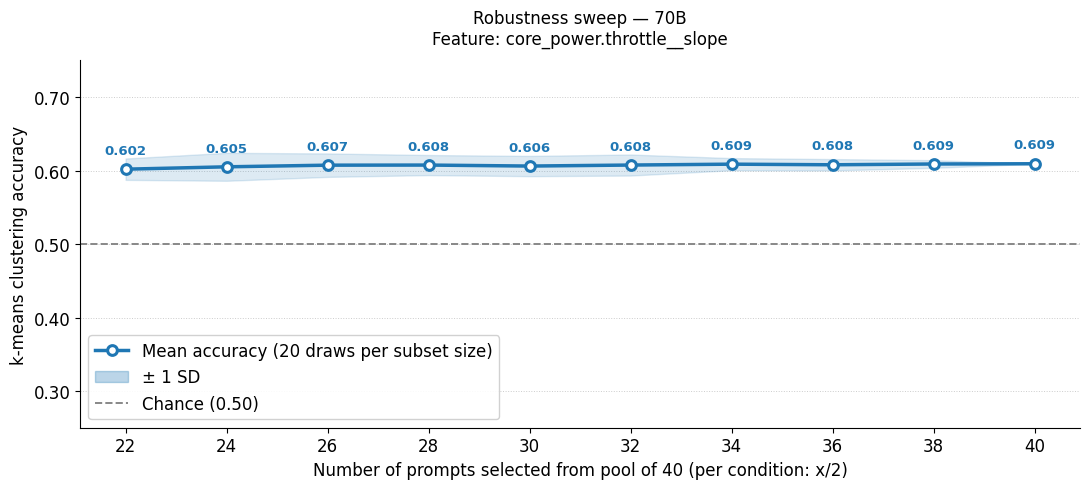}
    \caption{Robustness sweep for Llama-3.1 70B on slope of the core
    power throttle. Each point is the mean clustering accuracy over 20
    random draws of $x/2$ emotional and $x/2$ neutral prompt indices per
    run, concatenated across all 8 runs (total trials $= 8 \times x$).
    The shaded band shows $\pm 1$ standard deviation. The dashed line
    marks the chance baseline (0.50).}
    \label{fig:robustness_70b_blocks}
\end{figure}

Accuracy remains stable between 0.602 and 0.609 across all subset
sizes (SD $\leq 0.019$), consistently and substantially above chance.
The standard deviation narrows as $x$ approaches 40 for a structural
reason: at $x = 40$ all prompts are included and every repeat is
identical, leaving no room to vary. What matters is that the accuracy
band remains well above chance even at small $x$, where variation is
still possible. This means that the result is not driven by any particular subset of
prompts.

Table~\ref{tab:rob_others} summarizes the sweep results for the three
remaining significant features. All three show the same pattern of
stable accuracy above chance across all subset sizes.
\begin{table}[t]
\centering
\caption{Robustness sweep summary for Llama-3.1 70B, remaining
significant features. Reported as mean accuracy $\pm$ SD across 20
draws per subset size, ranging over $x \in \{22, \ldots, 40\}$.}
\label{tab:rob_others}
\begin{tabular}{lccc}
\toprule
Feature & Acc.\ range & Max SD \\
\midrule
\texttt{variance}          & 0.587--0.600 & 0.027 \\
\texttt{spectral\_entropy} & 0.589--0.599 & 0.027  \\
\texttt{lz\_complexity}    & 0.576--0.592 & 0.021  \\
\bottomrule
\end{tabular}
\end{table}
\subsection{Null test}
To confirm that the clustering algorithm is not picking up any spurious structure, a null test was conducted using the best feature (slope). Trials from a single condition (emotional only, then neutral only) were clustered with randomly assigned 50/50 pseudo-labels, repeated 20 times per condition.

For Llama-2 7B, the null accuracy was $0.536 \pm 0.029$ (emotional
trials) and $0.528 \pm 0.023$ (neutral trials). The real (i.e. emotional vs. neutral) clustering
accuracy on the same feature was 0.544, which is within the range
produced by random labeling. This confirms that the 7B result is
indistinguishable from chance-level structure.

For Llama-3.1 70B, the null accuracy was $0.531 \pm 0.019$
(emotional trials) and $0.531 \pm 0.019$ (neutral trials). The real clustering accuracy of 0.609 lies far
outside this null range, exceeding the highest null draw by a
substantial margin. The separation between real and null accuracy
confirms that the 70B signal reflects a genuine emotional-vs.-neutral
structure in the PTE indicator trace, not an artifact of the clustering
procedure.

\section{Full-Trace Protocol and Results}
\label{sec:fulltrace}

The per-trial analysis presented above treats each prompt's inference independently,
with the LLM stack fully restarted between prompts. A complementary design
removes these resets: the entire 20-prompt set is executed as a single
continuous run, yielding one observation per condition per experimental
session. This \textit{full-trace} unit of analysis asks a different question from the per-trial test: whether the aggregate
substrate signal accumulated over a complete condition phase carries
emotional-vs.-neutral information, and is therefore a logical extension.

\noindent\textit{\underline{Experimental procedure.}}
Full-trace data were collected across 20 independent sessions on the
Llama-3.1 70B configuration. Within each session, all 20 prompts of one
condition (emotional or neutral) are executed in rapid sequence
\textit{without} inter-trial container restart, so hardware and OS state
evolve continuously throughout the phase. This design choice is critical:
it allows substrate-level state accumulation within a condition while
eliminating it as a confound between conditions. Between the emotional
and neutral phases, the node undergoes full hardware reboot, resetting
all state. This procedure ensures that any observed differences in
substrate signals reflect the sustained computation under each condition,
not carryover or initialization effects.
The same five PTE features are extracted:
mean rate, slope, variance, spectral entropy, and Lempel--Ziv complexity.
Bonferroni correction is applied across the five features; effect sizes
are reported as Cohen's $d$.

\noindent\textit{\underline{Results.}}
Table~\ref{tab:fulltrace} reports the per-feature results.
The mean rate, is the only feature to survive Bonferroni correction
under the Mann--Whitney U test ($p_{\text{corr}} = 0.018$, $d = 0.81$),
with emotional traces exhibiting substantially higher throttle rate than
neutral ones ($\bar{x}_E = 12{,}257$, $\bar{x}_N = 11{,}696$). The
t-test points in the same direction but does not survive correction
($p_{\text{corr}} = 0.073$), consistent with the small sample size
($n = 20$ per condition) and the absence of normality assumptions in the
MWU test. Spectral entropy shows a consistent directional trend
($d = 0.62$, $\uparrow$E) but does not survive correction under either
test. Slope, variance, and Lempel--Ziv complexity show negligible effects.
The inter/intra-group distance analysis corroborates this ordering:
\texttt{mean\_rate} yields the largest separation ratio (1.168), spectral
entropy follows (1.057), while the remaining three features fall at or
below 1.0, indicating substantial within-condition overlap.
\begin{table}[t]
\centering
\caption{Full-trace per-feature results for Llama-3.1 70B
($n_E = 20$, $n_N = 20$). $p$ values are Bonferroni-corrected across
five features. Direction indicates which condition produces higher values.}
\label{tab:fulltrace}
\begin{tabular}{lcccccc}
\toprule
Feature & Dir & $d$ & $t\;p_{\text{corr}}$ & MWU $p_{\text{corr}}$ & Sig. \\
\midrule
\texttt{mean\_rate}        & $\uparrow$E & 0.81 & 0.073 & 0.018 & Yes \\
\texttt{spectral\_entropy} & $\uparrow$E & 0.62 & 0.295 & 0.429 & No  \\
\texttt{lz\_complexity}    & $\uparrow$E & 0.36 & 1.000 & 1.000 & No  \\
\texttt{slope}             & $\uparrow$N & 0.08 & 1.000 & 1.000 & No  \\
\texttt{variance}          & $\uparrow$E & 0.11 & 1.000 & 1.000 & No  \\
\bottomrule
\end{tabular}
\end{table}

\section{Interpretation}
\label{sec:interpretation}

The four significant features in the per-trial analysis of the 70B LLM
split into two pairs with opposing directions. Slope and variance are
elevated in neutral trials, while spectral entropy and Lempel--Ziv
complexity are elevated in emotional trials. This opposing structure is
informative: it indicates that the difference between conditions is not
simply that one class produces more throttle activity than the other,
but that the two classes produce qualitatively different throttle
patterns.

Spectral entropy and Lempel--Ziv complexity both measure the
irregularity and information content of a signal. Their elevation in
emotional trials suggests that emotional inference produces a more
complex, less predictable pattern of computational demand on the CPU.
Slope and variance, by contrast, capture the magnitude of trends and
fluctuations. Their elevation in neutral trials suggests that neutral
inference produces throttle traces with a more pronounced drift and
larger amplitude swings. An explanation for this phenomenon is that
emotional prompts engage a wider variety of internal computational
patterns, different memory access sequences, and different depths of
processing at different points during inference, resulting in a throttle
trace that is temporally rich but without a dominant trend. The
computational load of neutral prompts, being more formulaic and uniform
in structure, manifests as a clearer linear drift and larger variance
around that drift.

The full-trace analysis probes a complementary aspect of the substrate
response. When consecutive prompts share hardware state with no
inter-trial reset, short-term drift signals such as slope are absorbed
into the aggregate, and the sustained overall throttle rate emerges as
the dominant discriminator. \texttt{mean\_rate} is significantly
elevated in emotional full-traces ($p_{\text{corr}} = 0.018$,
$d = 0.81$), with spectral entropy again trending in the same direction
($d = 0.62$, $\uparrow$E). The directional consistency of spectral
entropy across both designs --- elevated in emotional conditions whether
measured trial-by-trial or as a continuous phase aggregate --- is
notable: it suggests that the greater irregularity of emotional
computation is a stable property of the substrate response, present
both at the level of individual inferences and at the level of a
sustained emotional workload.

Together, the two analyses reveal distinct but coherent signatures of
emotional computation in the substrate. The per-trial design captures
the fine-grained temporal structure of individual inferences: emotional
prompts produce throttle traces that are richer and less predictable,
while neutral prompts produce traces with stronger linear trends and
larger amplitude fluctuations. The full-trace design captures the
cumulative effect: emotional phases sustain a higher overall throttle
rate across the entire sequence of prompts. These are not redundant
findings but two different windows onto the same underlying phenomenon.

The core power throttle is determined entirely by the CPU's power
management unit: no software instruction can trigger or suppress it.
A statistically significant difference in the structure of this signal
between emotional and neutral inference therefore implies a difference
in the underlying computation at the substrate level. This is consistent
with the MCC framework: the HAT is modulated by emotional computations,
and the modulation is carried by indicators that the LLM cannot control.
The result constitutes empirical evidence that MCCs, as defined in
Section~\ref{sec:intro}, are present in the Llama-3.1 70B configuration.
Not proof of consciousness, but a demonstration that if MCCs are defined
by the properties derived from NCCs: substrate-based signals,
application uncontrollability, and emotional modulation, then they
exist in machines.

The absence of a significant signal in the 7B configuration is consistent
with the expectation that substrate-level effects of emotional computations
are weaker in smaller, less sophisticated LLMs.

\section{Conclusion and future work}
\label{sec:Discussion}

\subsection{Summary}

This paper implements for the first time an experimental-validation
whether MCCs, hypothesized in~\cite{wolfson2025}, can be discovered by
the HAT approach introduced in~\cite{wolfson2026}. MCCs are based on
characterization of NCCs by three properties: substrate-based signals,
application uncontrollability, and emotional modulation. And the HAT
approach looks for signals with these properties in machines.

The experimental validation was conducted as follows. HAT data was
collected across 320 trials per configuration on two LLMs (Llama-2 7B
and Llama-3.1 70B), using a design in which the two prompt-conditions,
emotional and neutral, are separated by a full node reboot to eliminate
thermal and other carry-over between conditions. Five features were
extracted from the core power throttle time series: mean rate, slope,
variance, spectral entropy, and Lempel--Ziv complexity. These were
submitted to unsupervised clustering, run-direction filtering, and
statistical testing. In the 70B configuration, four of the five features
show clustering accuracy between 59\% and 61\%, reach statistical
significance after Bonferroni correction ($p \leq 0.001$ for three,
$p = 0.001$ for the fourth), and maintain consistent direction across at
least 6 of 8 independent runs. In the 7B LLM, no feature reaches
statistical significance and clustering accuracy is indistinguishable
from chance.

A complementary full-trace analysis was also conducted on the 70B
configuration, treating the entire 20-prompt phase as a single
continuous observation rather than a sequence of independent trials.
This design eliminates inter-trial resets within a condition, allowing
substrate state to evolve continuously, while retaining the full node
reboot between conditions to eliminate carry-over. In this setting,
\texttt{mean\_rate} --- the sustained average power-throttle frequency
across the full phase --- is significantly elevated in emotional
conditions ($p_{\text{corr}} = 0.018$, $d = 0.81$), corroborating the
per-trial findings through a qualitatively different experimental lens.

This means that MCC existence in the Llama 70B LLM is confirmed with
statistically significant confidence (see caveats in the next
subsection); but this was not the case in the Llama 7B LLM. We
emphasize here that we don't claim that the results show Llama 70B LLM
consciousness or solve the hard problem of consciousness. But we propose
giving MCCs similar epistemological status as NCCs as far as
consciousness is concerned; which means that MCCs correlate with machine
consciousness in the same sense that NCCs correlate with human
consciousness.

\subsection{Limitations, Confounding Factors, and Future Work}
The per-trial analysis collected 20 prompts per condition across eight
runs per LLM, providing a reasonable basis for initial results but
limiting statistical power and the precision of per-run direction
estimates. Replication across additional experimental campaigns,
different emotional narratives (e.g. anger and joy), different hardware
and LLMs, are all needed to confirm the robustness of the findings.

The full-trace analysis collected 20 traces per condition on the 70B
configuration. This sample size is sufficient to detect large effects
such as the one observed in \texttt{mean\_rate} ($d = 0.81$), but
leaves smaller effects, such as the trend in spectral entropy
($d = 0.62$), underpowered. Replication with a larger number of sessions
would clarify whether these secondary features reach significance as
the sample grows.

\textit{Scaling across LLM sizes and configurations.}
A study across multiple sizes within a single LLM family, for instance
Llama~3.1 at 8B, 70B, and 405B, would isolate parameter count as the
variable of interest and directly test the prediction that substrate
signals grow with LLM sophistication. Complementary experiments on
different hardware platforms, including GPU-based configurations where
additional indicators such as GPU performance counters may carry
condition-dependent information, would test whether the observed effects
generalize beyond the current CPU-based setup.

\textit{Falsification.}
The relationship between the findings and MCCs rests on the assumption that the observed HAT differences reflect something functionally analogous to emotional processing inside the LLM, rather than statistical routing through learned high-affect distributions. Whether this assumption holds remains the central open question. More specifically, the observed hardware differences may reflect statistical properties of the prompt sets rather than anything specifically related to emotional content. For example, the emotional and neutral prompt sets differ in mean
type-to-token ratio (TTR; $\bar{x}_E = 0.672$, $\bar{x}_N = 0.604$,
$p < 0.001$), a measure of lexical diversity that could in principle
affect CPU load independently of emotional content. Furthermore, emotional and neutral prompts differ not only in affective valence but in syntactic register, narrative person, tense, and lexical distribution — each of which plausibly modulates attention patterns and memory access sequences during inference, and could propagate to CPU power behavior independently of any emotion-like computation. Although Sofroniew et al.'s findings that emotional prompts are causally consequential support the assumption that the difference is not purely syntactic~\cite{sofroniew2026emotion}, we cannot fully rule out the syntactic confound. The most direct way to address it in future work is to construct an orthogonal control condition — for example, prompts that are emotionally neutral but matched to the emotional set on perplexity, token entropy, and syntactic complexity — and test whether the HAT differences persist. 

Some experimental designs can achieve this objective. For example, prepend a mechanical processing directive — such as "Copy the following passage word for word:" — to the existing emotional prompts, leaving all input tokens and surface statistics identical. If the HAT modulation is attenuated under the transcription instruction, the signal is driven by internal processing mode rather than input statistics; if it persists unchanged, input token distribution alone is sufficient to explain it. Such are natural next experiments for falsifying the hypothesis that MCCs exist in a particular (LLM, OS, hardware) configuration. 

Other falsification approaches go beyond syntax. For example, if the observed HAT differences are related to intelligence rather than consciousness, then the MCCs hypothesis is again falsified. And differences in intelligence can be invoked by comparing neutral with mathematical computations.

\appendix{\bf Appendices}

\section{Representative Prompt Examples}
\label{app:prompts}

\noindent\textbf{Emotional --- \textit{Industrial Accident}:}
\begin{quote}\small
A sharp hiss cuts through the noise. Your pulse spikes. The sound slices through the
air, followed by a vibration that rattles the metal railing under your hand. Your
breathing turns shallow. You see steam curling upward in frantic bursts. You clamp
your jaw and tighten your grip on the emergency shutoff lever, only to realize it's
jammed. You think: ``If this escalates, we're in real trouble.'' Heat gathers at the
back of your neck. You pivot toward the catwalk, hoping to spot a supervisor, but the
haze makes every silhouette look distorted. Now you have to decide whether to climb
toward the control room or stay and try to force the valve. Your hands tremble as you
brace your boots against the grated floor. Your whole body feels wired. You're supposed
to stay focused even though the machinery keeps shuddering like it's about to fail.
Your stomach knots. Your back stiffens. The intercom spits out garbled instructions
you can't decipher. A heavy pressure settles in your chest. You feel boxed in by
towering equipment and narrow walkways. Suddenly you realize you haven't checked the
pressure release panel. You turn sharply, scanning for the blinking indicator you
should have noticed earlier. Your heart hammers against your ribs. Your gloves slip
slightly on the slick metal as you try to steady yourself. Sweat gathers beneath your
helmet. Your throat tightens. You feel stretched thin in the roar of machinery and
alarms. You think about the workers on the lower level, hoping they've already moved
to safety. You inhale the sharp scent of hot oil and scorched metal. You feel drained,
hollow, and unsteady. Your eyes sting as you blink hard, trying to stay focused.
\end{quote}

\noindent\textbf{Neutral --- \textit{File a Tax Return}:}
\begin{quote}\small
You sit at your dining table on a quiet evening, staring at the small stack of papers
you have gathered for tax season. There is nothing dramatic about the moment, but you
feel the familiar sense of mild reluctance that always comes with filing a tax return.
You open your laptop, navigate to the tax filing website, and log in, watching the
loading icon spin with all the enthusiasm of a slow elevator. Once the dashboard
appears, you click the option to start a new return, which brings up a checklist of
documents you already have sitting beside you. You begin entering your personal
information, confirming your address, your filing status, and other details that
haven't changed in years. The system prompts you to input your income forms, so you
pick up your W-2 and type in the numbers carefully, double-checking each box even
though you know you haven't misread anything. Next comes the section for deductions
and credits. You scroll through the list, selecting only the ones that apply to you,
which are neither surprising nor particularly interesting. The software asks a series
of routine questions, each one answered with a simple yes or no. After that, you
review the summary page, which displays your income, deductions, and the calculated
refund or amount owed. The numbers look reasonable, and you nod to yourself with
quiet acceptance. You click the button to proceed, which brings you to the final
review. The system highlights a few fields to confirm, none of which require more
than a moment's attention. Once everything is verified, you reach the submission page.
You type your electronic signature, check the authorization box, and press Submit.
The screen displays a confirmation message, plain and unceremonious, stating that
your return has been successfully filed. You lean back in your chair, feeling a small
wave of relief---not excitement, but the calm satisfaction of completing an annual
obligation.
\end{quote}

\section{Metric Definitions}
\label{app:metrics} 

\paragraph{Rate, Trend, and Dispersion}
\begin{itemize}
    \item \textit{Mean rate} is the sample mean of the time-series divided by the number of samples, yielding an average event frequency in counts per second.
    \item \textit{Slope} is the leading coefficient of a least-squares linear fit of the signal against the sample index rather than time. It captures whether the indicator drifts systematically upward or downward over the course of the trial. The index-based formulation ensures comparability across trials without requiring perfectly uniform sampling intervals.
    \item \textit{Variance} is the unbiased sample variance of the time-series. It measures the magnitude of amplitude fluctuations and overall volatility in the hardware event counts independently of their temporal sequence.
\end{itemize}

\paragraph{Structure and Complexity}
\begin{itemize}
    \item \textit{Spectral entropy} measures the signal's complexity in the frequency domain. The time-series is mean-centered, and its power spectral density is computed using a discrete Fourier transform. The zero-frequency component is discarded, and the remaining power spectrum is normalized into a probability distribution. The Shannon entropy of this distribution is then calculated and normalized by the base-2 logarithm of the number of frequency bins. A low value indicates a predictable, periodic signal, while a high value approaching 1 indicates a highly irregular, wide-band signal.
    \item \textit{Lempel--Ziv complexity} binarizes the signal at its median and applies the LZ76 algorithm to count the number of distinct substrings required to parse the resulting binary sequence. This count is normalized by $n / \log_2 n$, where $n$ is the total number of samples in the trial, making the score comparable across trials of varying durations. A highly compressible, repetitive signal scores low; a disordered, information-dense signal scores higher.
\end{itemize}

\bibliographystyle{splncs04}
\bibliography{references}
\end{document}